\documentclass[10pt,twocolumn,letterpaper]{article}

\usepackage{iccv}
\usepackage{times}
\usepackage{epsfig}
\usepackage{graphicx}
\usepackage{amsmath}
\usepackage{amssymb}

\usepackage{caption}
\usepackage{subcaption}
\usepackage{tablefootnote}

\newcommand{\Mark}[1]{\textsuperscript#1}

\usepackage[pagebackref=true,breaklinks=true,letterpaper=true,colorlinks,bookmarks=false]{hyperref}

\iccvfinalcopy 

\def\iccvPaperID{1375} 

\ificcvfinal\fi

\begin{document}

\title{Learning to Disambiguate by Asking Discriminative Questions}

\author{
	\makebox[\linewidth][c]{Yining Li\Mark{1}\hspace{1.5em}Chen Huang\Mark{2}\hspace{1.5em}Xiaoou Tang\Mark{1}\hspace{1.5em}Chen Change Loy\Mark{1}}\\
	\Mark{1}Department of Information Engineering, The Chinese University of Hong Kong\\
	\Mark{2}Robotics Institute, Carnegie Mellon University\\
	{\ttfamily\small \{ly015, xtang, ccloy\}@ie.cuhk.edu.hk, chenh2@andrew.cmu.edu}
}

\maketitle

\begin{abstract}
The ability to ask questions is a powerful tool to gather information in order to learn about the world and resolve ambiguities. 
In this paper, we explore a novel problem of generating discriminative questions to help disambiguate visual instances. 
Our work can be seen as a complement and new extension to the rich research studies on image captioning and question answering.
We introduce the first large-scale dataset with over 10,000 carefully annotated images-question tuples to facilitate benchmarking. In particular, each tuple consists of a pair of images and 4.6 discriminative questions (as positive samples) and 5.9 non-discriminative questions (as negative samples) on average.
In addition, we present an effective method for visual discriminative question generation. The method can be trained in a weakly supervised manner without discriminative images-question tuples but just existing visual question answering datasets. Promising results are shown against representative baselines through quantitative evaluations and user studies.

\end{abstract}

\section{Introduction}
\label{sec:introduction}

\begin{figure}[t]
\centering
\includegraphics[width=\linewidth]{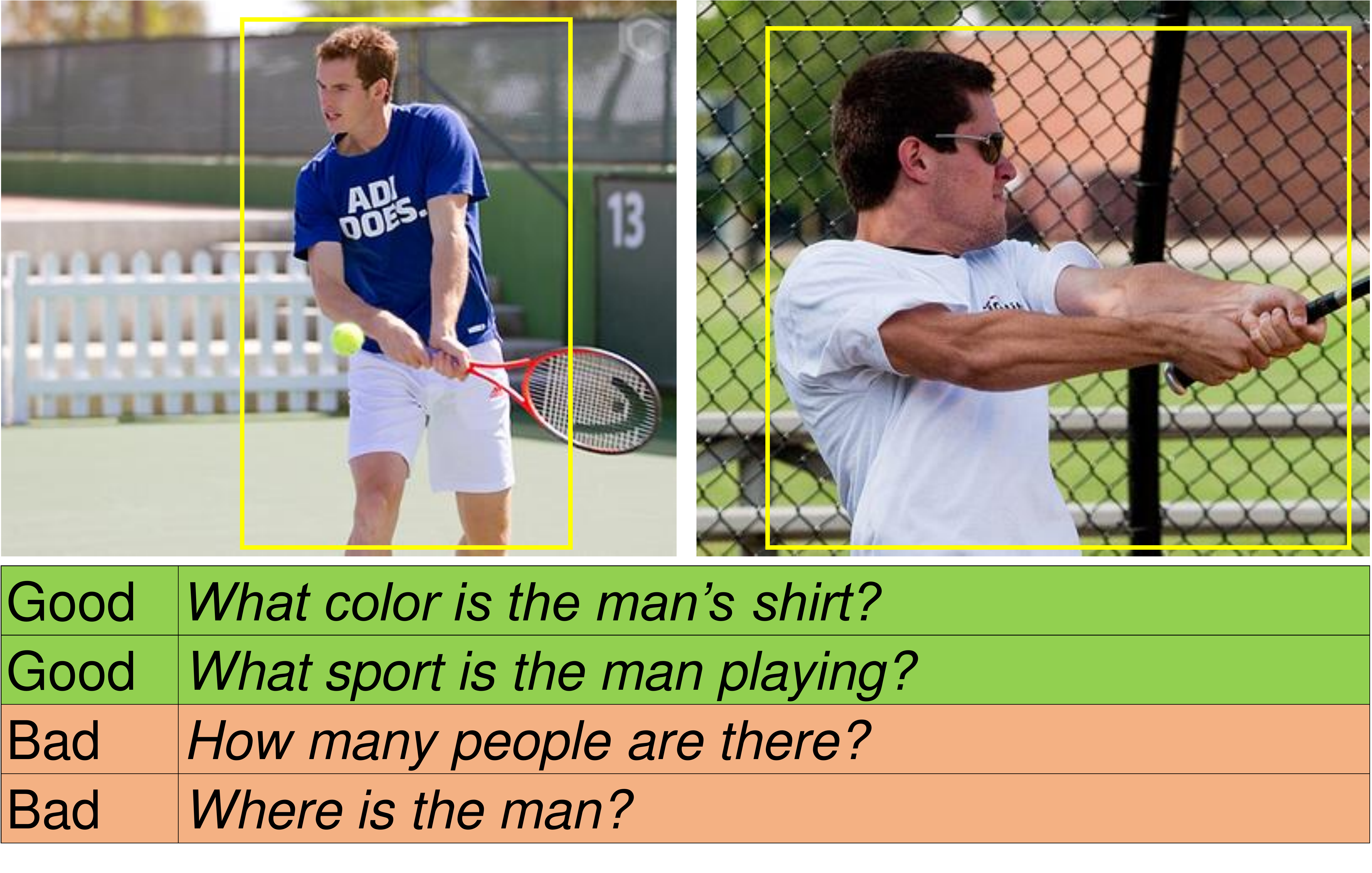}
\vskip -0.2cm
\caption{Example ambiguous image pair and both good and bad discriminative questions.}
\label{fig:intro}
\vskip -0.2cm
\end{figure}

Imagine a natural language dialog between a computer and a human (see Fig.~\ref{fig:intro}):

Kid\hspace{0.95cm}: \textit{``What sport is the man playing?''.}

Computer\hspace{0.08cm}: \textit{``What is the color of his shirt?''}

Kid\hspace{0.95cm}: \textit{``Blue.''}

Computer\hspace{0.08cm}: \textit{``He is playing tennis.''}

\noindent In the conversation, the kid refers to an object but the description is linguistically underspecified. The computer analyzes the visual differences of plausible targets and then reacts by asking a discriminative question \textit{``What is the color of his shirt?''} to resolve the reference.

We define the aforementioned problem as \textit{Visual Discriminative Question Generation} (VDQG). Specifically, the computer is given with two visual instances, and the goal is to ask a good question to distinguish or disambiguate them. In this study, we call the pair images as ambiguous pairs -- the ambiguity may not necessarily be due to their subtle visual differences. They may just belong to the same object class with close proximity in their deep representation. Although such ambiguity can be easily resolved by human, they can be difficult to a machine.
Distinguishing different image pairs require asking different types of questions, ranging from color, action, location, and number.
Akin to the classic ``Twenty Questions'' game, a careful selection of questions can greatly improve the odds of the questioner to narrow down the answer. A bad question would fail to eliminate ambiguities. Figure~\ref{fig:intro} gives good and bad examples of questions.
This questioning capability can subsequently be extended to generating a sequence of discriminative questions and prompting a human-in-the-loop to answer them. In the process, the machine accumulates evidence that can gradually refine the language expression from humans and finally distinguish the object of interest.

Such VDQG ability allows a machine to play a more natural and interactive role in Human-Computer Interaction (HCI), or improve a robot to bind the references made by a speaker more accurately to objects in a scene.
While there have been various attempts to build a system that can provide explanations~\cite{hendricks2016generating} or ask questions~\cite{mora2016towards,mostafazadeh-EtAl:2016:P16-1} based on visual instances, the problem of VDQG has not been explored.
The goal of VDQG is to resolve inter-object ambiguities through asking questions. It is thus differs from image captioning that aims at generating a literal description based on a single visual instance. It also differs from Visual Question Answering (VQA), which takes an image and a question as inputs and provides an answer.
A closer work is Visual Question Generation (VQG)~\cite{mora2016towards,mostafazadeh-EtAl:2016:P16-1}. Unlike the setting of generating one possible question from an image, VDQG operates on two visual instances and generates a discriminating question for them. 
The most relevant work to ours is Yu~\etal~\cite{yu2016modeling}, which generates unambiguous referring expressions for an object by incorporating visual comparison to other objects in an image. Our problem differs in that we generate one single question to distinguish multiple objects instead of referring expressions for all objects.

It is non-trivial to train a machine to ask discriminative questions in an automatic and human understandable way. Firstly, it should ask a natural and object-focused question.
Secondly, and importantly, the machine is required to pinpoint the most distinguishing characteristics of two objects to perform a comparison.  
Addressing the problem is further compounded by the lack of data. In particular, there are no existing datasets that come readily with pair images annotated with discriminative questions. Thus we cannot perform a direct supervised learning.

To overcome the challenges, we utilize the Long Short-Term Memory (LSTM)~\cite{greff2015lstm} network to generate natural language questions. 
To generate discriminative questions, which are object-focus, we condition the LSTM with a visual deep convolutional network that predicts fine-grained attributes. 
Here visual attributes provide a tight constraint on the large space of possible questions that can be generated from the LSTM. We propose a new method to identify the most discriminative attributes from noisy attribute detections on the two considered objects. Then we feed the chosen attributes into the LSTM network, which is trained end-to-end to generate an unambiguous question.
To address the training data problem, we introduce a novel approach to training the LSTM in a weakly-supervised manner with rich visual questioning information extracted from the Visual Genome dataset~\cite{krishna2017visual}. In addition, a large-scale VDQG dataset is proposed for evaluation purposes.

\vspace{0.1cm}
\noindent\textbf{Contributions}: 
We present the first attempt to address the novel problem of Visual Discriminative Question Generation (VDQG). To facilitate future benchmarking, we extend the current Visual Genome dataset~\cite{krishna2017visual} by establishing a large-scale VDQG dataset of over $10,000$ image pairs with over $100,000$ discriminative and non-discriminative questions. 
We further demonstrate an effective LSTM-based method for discriminative question generation. Unlike existing image captioning and VQG methods, the proposed LSTM is conditioned on discriminative attributes selected through a discriminative score function.
We conduct both quantitative and user studies to validate the effectiveness of our approach.

\section{Related Work}
\label{sec:related_work}

\noindent
\textbf{Image Captioning}. The goal of image captioning is to automatically generate natural language description of images~\cite{fang2015captions}. The CNN-LSTM framework has been commonly adopted and shows good performance~\cite{donahue2015long,karpathy2015deep,mao2014deep,vinyals2015show,wu2017image}. Xu \etal~\cite{xu2015show} introduce attention mechanism to exploit spatial information from image context. Krishna \etal~\cite{krishna2017visual} incorporate object detection~\cite{ren2015faster} to generate descriptions for dense regions. Jia \etal~\cite{jia2015guiding} extracts semantic information from images as extra guide to caption generation. Krause \etal~\cite{krause2016hierarchical} uses hierarchical RNN to generates entire paragraphs to describe images, which is more descriptive than single sentence caption.
In contrast to these studies, we are interested in generating a question rather than a caption to distinguish two objects in images.

\noindent
\textbf{Visual Question Answering (VQA)}. 

VQA aims at generating answer given an input image and question. It differs from our task of generating questions to disambiguate images. Deep encoder-decoder framework~\cite{malinowski2015ask} has been adopted to learn a joint representation of input visual and textual information for answer prediction (multiple-choice) or generation (open-ended). Visual attention~\cite{lu2016hierarchical,shih2016look,xu2016ask,yang2016stacked} and question conditioned model~\cite{andreas2016neural, noh2016image} have been explored to capture most answer-related information from images and questions. To facilitate VQA research, a number of benchmarks has been introduced~\cite{antol2015vqa,krishna2017visual, mostafazadeh2017image, ren2015exploring, yu2015visual, zhu2016visual7w}. Johnson \etal~\cite{johnson2016clevr} introduce a diagnostic VQA dataset by mitigating the answer biases which can be exploit to achieve inflated performance. Das \etal~\cite{das2016visual} extend VQA to a dialog scenario. Zhang \etal~\cite{zhang2016yin} build a balanced binary VQA dataset on abstract scenes by collect counterpart images that yield opposite answers to the same question. A concurrent work to ours is~\cite{goyal2016making}, which extends the popular VQA dataset~\cite{antol2015vqa} by collecting complementary images such that each question will be associated to a pair of similar images that result in to different answers. Both \cite{zhang2016yin} and \cite{goyal2016making} contribute a balanced VQA dataset do not explore the VDQG problem. Although our model can be trained on balanced VQA data, we show that it performs reasonably well by just learning from unbalanced VQA datasets.

\noindent
\textbf{Referring Expression Generation (REG)}. A closely related task to VDQG is REG, where the model is required to generate unambiguous object descriptions. Referring expression has been studied in Natural Language Processing (NLP)~\cite{golland2010game, krahmer2012computational,winograd1972understanding}. Kazemzadeh ~\etal~\cite{kazemzadeh2014referitgame} introduce the first large-scale dataset for the REG in real-world scenes. They use images from the ImageCLEF dataset~\cite{escalante2010segmented}, and collect referring expression annotations by developing a ReferIt game. 
%
%
The authors of~\cite{mao2016generation,yu2016modeling} build two larger REG datasets by using similar approaches on top of MS COCO~\cite{lin2014microsoft}. CNN-LSTM model has been shown effective in both generation~\cite{mao2016generation,yu2016modeling} and comprehension~\cite{hu2016natural, nagaraja2016modeling} of REG. Mao \etal~\cite{mao2016generation} introduce a discriminative loss function based on Maximum Mutual Information. Yu \etal~\cite{yu2016modeling} study the usage of context in REG task. Yu \etal~\cite{yu2016joint} propose a speaker-listener-reinforcer framework for REG, which is end-to-end trainable by reinforcement learning.

\begin{figure*}[t]
	\includegraphics[width=\linewidth]{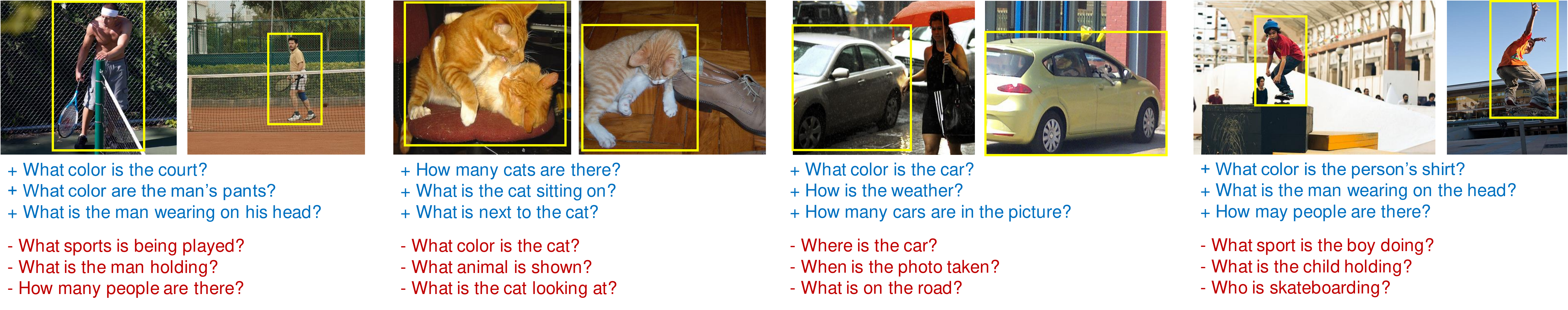}
	\vskip -0.25cm
	\caption{Example of ambiguous pairs and the associated positive and negative question annotations in the proposed VDQG dataset. Positive and negative questions are written in blue and red, respectively. More examples in supplementary material. }
	\label{fig:dataset_examples}
\end{figure*}

\noindent
\textbf{Visual Question Generation (VQG)}. 
Natural-language question generation from text corpus has been studied for years~\cite{ ali2010automation,chen2009aist,kalady2010natural,serban2016generating}. 
The task of generating question about images, however, has not been extensively studied. A key problem is the uncertainty of the questions' query targets, which makes the question generation subjective and hard to evaluate. Masuda-Mora \etal~\cite{mora2016towards} design a question-answer pair generation framework, where a CNN-LSTM model is used to generate image-related questions, and a following LSTM will decode the hidden representation of the question into its answer. Their model is trained using VQA annotations~\cite{antol2015vqa}.
Mostafazadeh \etal~\cite{mostafazadeh-EtAl:2016:P16-1} introduce the first VQG dataset. Mostafazadeh \etal~\cite{mostafazadeh2017image} further extend the scenario to image-grounded conversation generation, where the model is repurposed for generating a sequence of questions and responses given image contexts. These tasks are essentially same as image captioning, because the goal is to model the joint distribution of image and language (questions), without explicitly considering the query target of the generated question.
A concurrent work~\cite{de2016guesswhat} proposes to use yes-no question sequences to locate unknown objects in images, and introduces a large-scale dataset. This work strengthens our belief on the importance of visual disambiguation by natural-language questions. The differences between this work and ours are: 1) We do not restrict a question to be yes-no type but more open-ended. 2) We explore the usage of semantic attributes in discriminative question generation. 3) No training data is available for training our VDQG. We circumvent this issue through a weakly-supervised learning method, which learns discriminative question generation from general VQA datasets.

\section{VDQG Dataset for Evaluation} 
\label{sec:dataset}

Existing VQG and VQA datasets \cite{antol2015vqa,johnson2016clevr, krishna2017visual,mostafazadeh-EtAl:2016:P16-1,ren2015exploring,zhu2016visual7w} only contain questions annotated on single image\footnote{Apart from the concurrent work~\cite{goyal2016making}, which released a large-scale balanced VQA dataset. Unfortunately the dataset was released in late March so we were not able to train/test our model on this data.}, which is inadequate for quantitative evaluation and analysis of VDQG methods. To fill the gap, we build a large-scale dataset that contains image pairs with human-annotated questions. We gather images from the Visual Genome dataset~\cite{krishna2017visual} and select image pairs as those that possess the same category label and high CNN feature similarity. Finally we employ crowd-sourcing to annotate discriminative and non-discriminative questions on these pairs. 
Some of the example pairs and the associated questions are shown in Fig.~\ref{fig:dataset_examples}. As can be observed, many of these pairs are ambiguous not only because they are of the same object class, but also due to their similar visual appearances.
We detail the data collection process as follows.

\noindent
\textbf{Ambiguous Pair Collection}. The Visual Genome dataset provides object annotations with their category labels and bounding boxes. We select 87 object categories that contain rich and diverse instances. 
Incorrect labeled and low-quality samples are discarded.
Subsequently, we cluster image instances in each object category by their features extracted with Inception-ResNet~\cite{szegedy2016inception}. Image pairs are randomly sampled from a cluster to form the ambiguous pairs.

\begin{table}[t]
	\centering
	\caption{Statistics of VDQG dataset. The length of a question is given by the number of tokens.}
	\label{table:dataset_statistics}
	\small{
	\begin{tabular}{l c}
		\hline
		No. of images & $8,058$\\
		No. of objects &  $13,987$\\
		No. of ambiguous image pairs & $11,202$\\
		No. of questions & $117,745$\\
		Avg. pos-question number per object pair & $4.57$\\
		Avg. neg-question number per object pair & $5.94$\\
		Avg. token number per question & $5.44$\\
		\hline
	\end{tabular}
	}
	\vskip -0.2cm
\end{table}

\noindent
\textbf{Question Annotation}. Question annotation is a laborious process. We therefore adopt a two-step approach to collect annotations by crowd-sourcing, and augment with more questions automatically followed by human verification.
In the first step, the workers are prompted to ask questions that can tell the differences between two images in an ambiguous pair. In this way we collect 2 to 3 discriminative questions for pair. It is worth pointing out that we collect `7W' questions, consistent with protocol adopted by the Visual Genome dataset~\cite{krishna2017visual}. This is the major difference between our dataset and ~\cite{de2016guesswhat}, which only contains `yes-no' questions. 

\begin{figure*}[t]
	\centering
	\begin{subfigure}[t]{0.32\textwidth}
		\includegraphics[width=\textwidth]{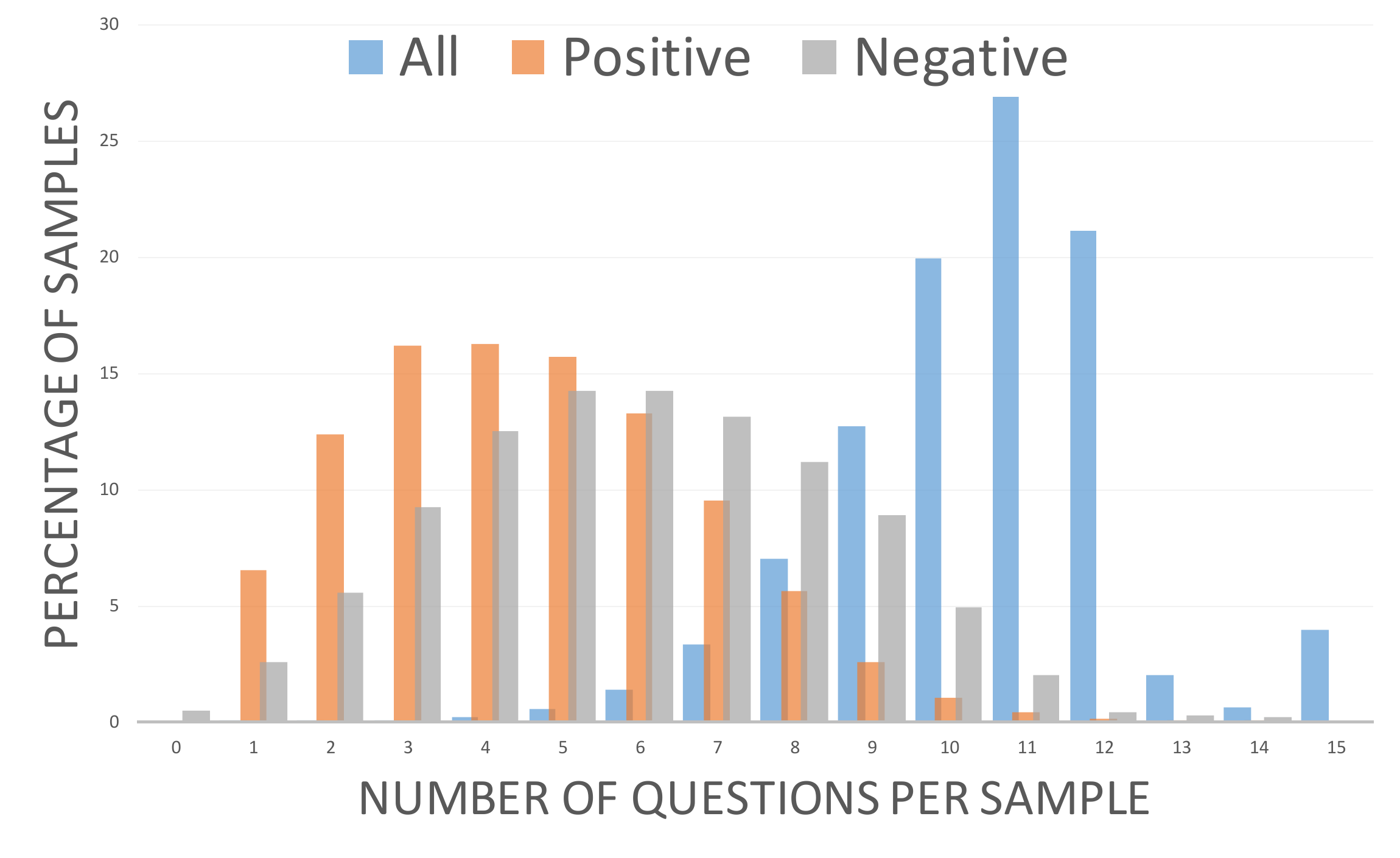}
		\caption{Distribution of number of questions}
		
	\end{subfigure}
	~
	\begin{subfigure}[t]{0.32\textwidth}
		\includegraphics[width=\textwidth]{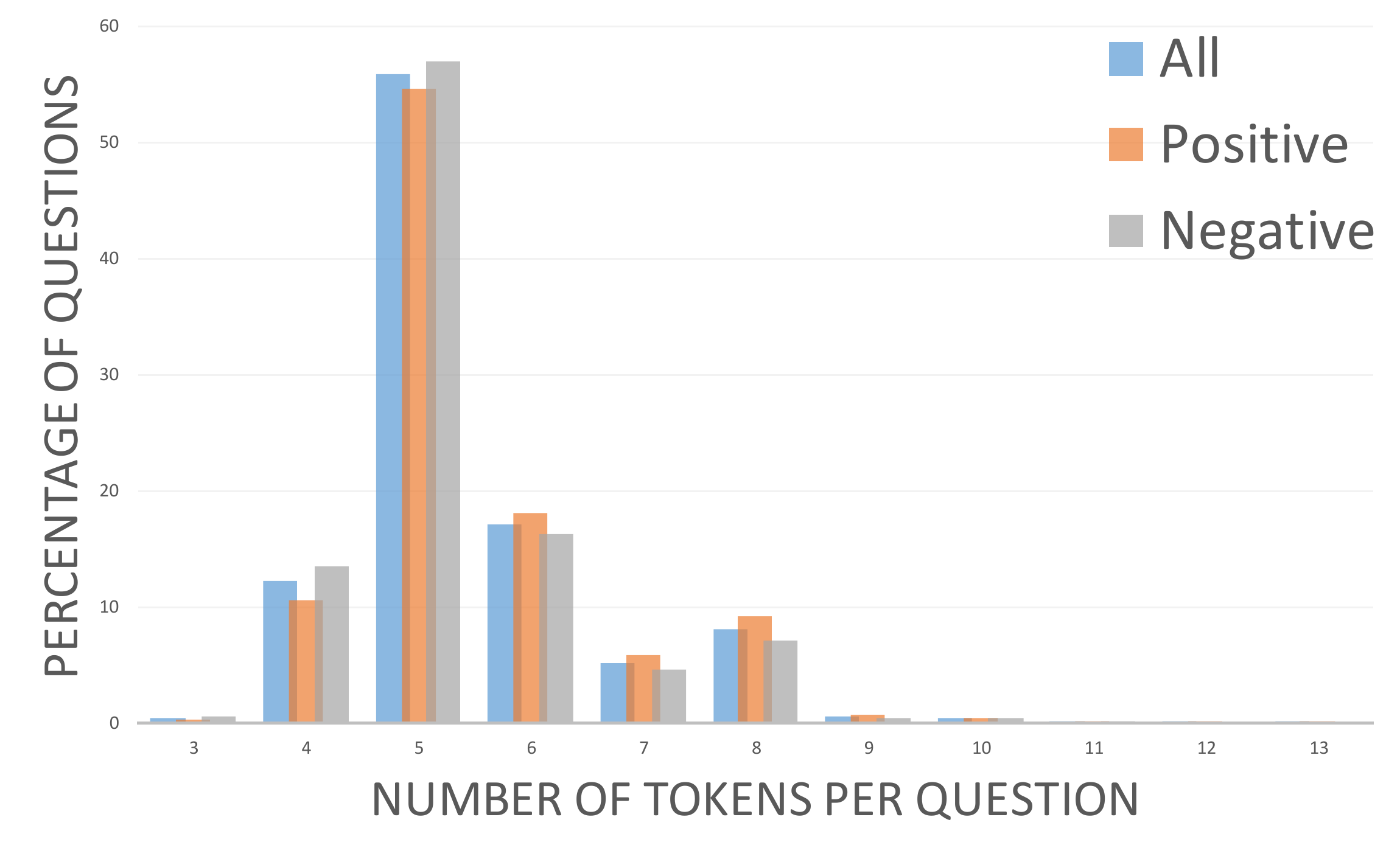}
		\caption{Distribution of question length}
		
	\end{subfigure}
	~
	\begin{subfigure}[t]{0.33\textwidth}
		\includegraphics[width=\textwidth]{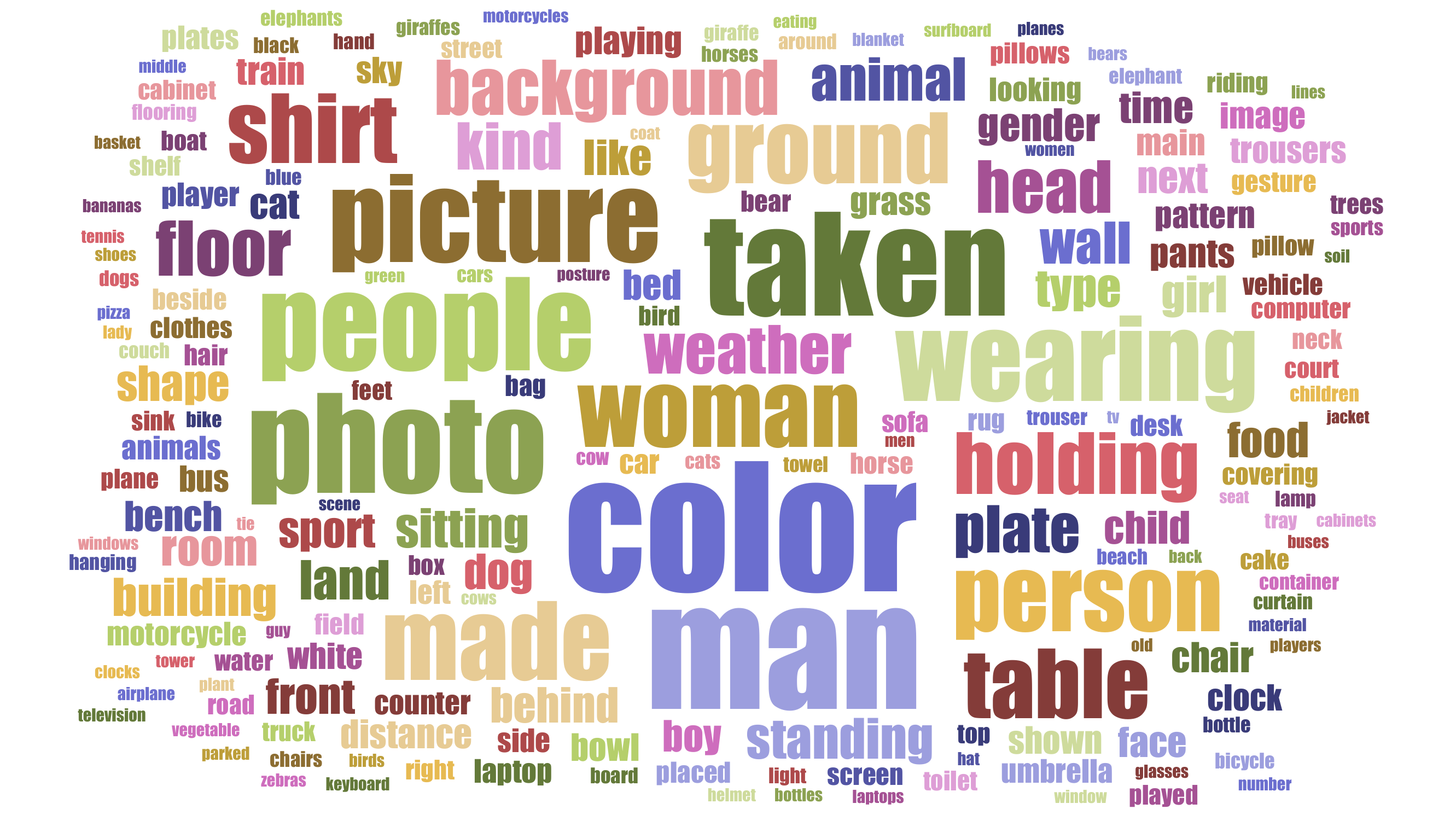}
		\caption{Word cloud of questions}
	\end{subfigure}
	\vskip -0.2cm
	\caption{Statistics of the VDQG dataset.}
	\label{fig:dataset_statistics}
	\vskip -0.2cm
\end{figure*}

Then we augment the question set of each ambiguous pair by 1) retrieving questions from other visually similar ambiguous pair and 2) automatically generating questions using a CNN-LSTM model trained on Visual Genome VQA annotations. 
After augmentation each ambiguous pair has over 8 question annotations. The added questions are expected to be related to the given images, but not guaranteed to be discriminative. Thus in the second step, the workers are shown with an ambiguous pair and a question, and they will judge whether the question would provide two different answers respectively to the images pair. Specifically, the worker will rate the question in a range of strong-positive, weak-positive and negative, which will serve as the label of the question.

\noindent
\textbf{Statistics}.
Our dataset contains $13,987$ images covering $87$ object categories. We annotated $11,202$ ambiguous image pairs with $117,745$ discriminative and non-discriminative questions. Table~\ref{table:dataset_statistics} summarizes key statistics of our dataset.
We provide an illustration in Fig.~\ref{fig:dataset_statistics} to show more statistics of the proposed dataset. Further statistics and examples of this dataset can be found in the supplementary material.

\section{Visual Discriminative Question Generation}
\label{sec:methodology}

\begin{figure*}[t]
	\centering
	\includegraphics[width=0.8\linewidth]{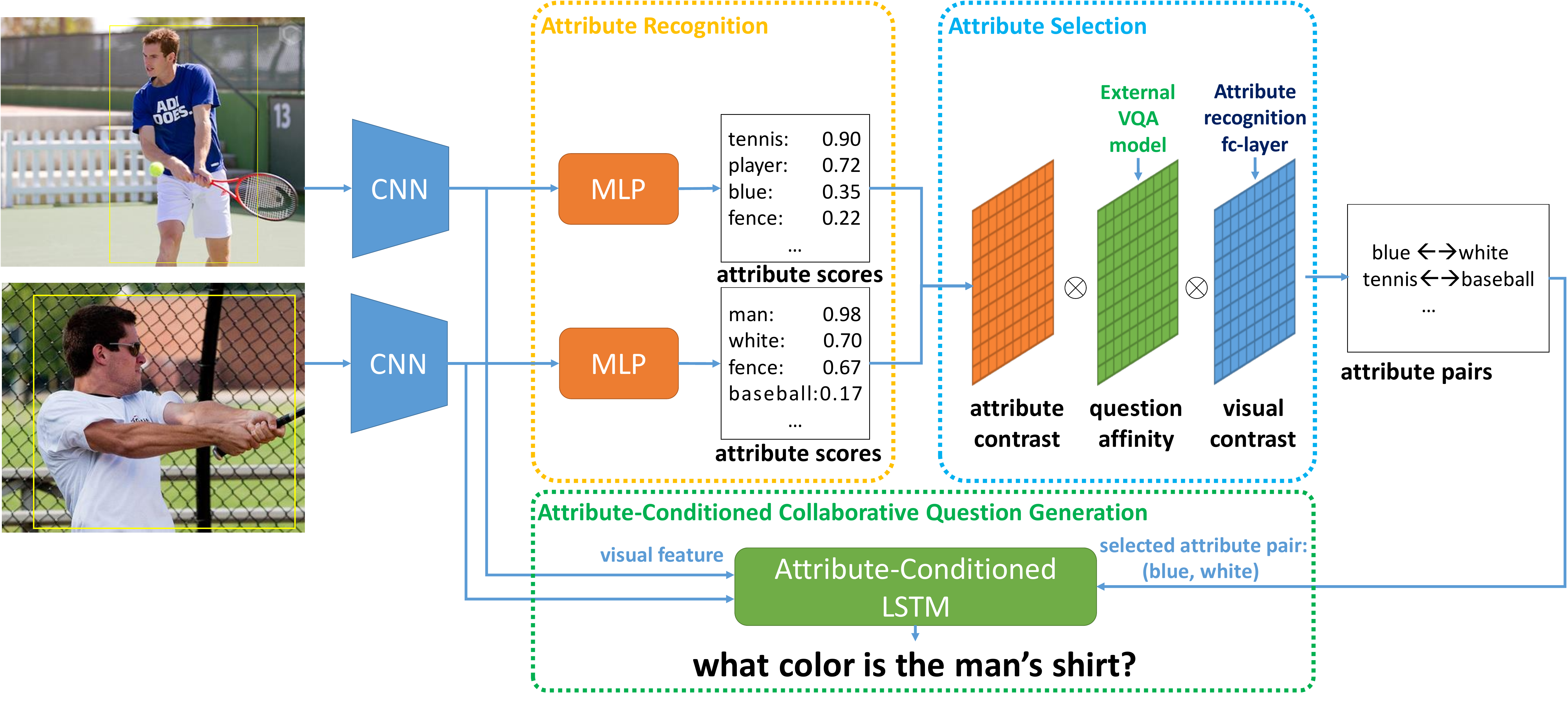}
	\vskip -0.25cm
	\caption{Overview of the attribute-conditioned question generation process. Given a pair of ambiguous images, we first extract semantic attributes from the images respectively. The attribute scores are sent into a selection model to select the distinguishing attributes pair, which reflects the most obvious difference between the ambiguous images. Then the visual feature and selected attribute pair are fed into an attribute-conditioned LSTM model to generate discriminative questions.}
	\label{fig:model}
	\vskip -0.25cm
\end{figure*}

Our goal is to generate discriminative questions collaboratively from two image regions $R^A$ and $R^B$. 
We show the proposed VDQG approach in Fig.~\ref{fig:model}.
The approach can be divided into two steps. The first step is to find discriminative attribute pairs. An attribute recognition and attribute selection components will be developed to achieve this goal. In particular, each region will be described by an attribute, and collectively, they should form a pair that best distinguish the two regions. For instance, as shown in Fig.~\ref{fig:model}, the `blue-white' attributes constitute a pair that is deemed more discriminative than the `tennis-baseball' pair, since the baseball bat is hardly visible. 
Given the discriminative attributes, the second step is to use the attributes to condition an LSTM to generate discriminative question. 

Inspired by~\cite{hu2016natural, mao2016generation}, the image region is represented by a concatenation of its local feature, image context and relative location/size: $\mathbf{f} = [f_{cnn}(R),f_{cnn}(I),\mathbf{l}_r]$. Specifically, $f_{cnn}(R)$ and $f_{cnn}(I)$ represent the 2048-d region and image features, respectively. The features are extracted using a Inception-ResNet~\cite{szegedy2016inception} pre-trained on ImageNet~\cite{russakovsky2015imagenet}. The vector $\mathbf{l}_r = [\frac{x_{tl}}{W},\frac{y_{tl}}{H},\frac{x_{br}}{W}, \frac{y_{br}}{H}, \frac{S_r}{S_I}]$ denotes the relative location and size of the region.

\subsection{Finding Discriminative Attribute Pairs}
\label{sec:attribute_model}

To find a pair of discriminative attributes, our method first recognizes visual attributes from each region to form an paired attribute pool. The method then applies attribute selection to select a pair of attributes that best distinguish the two regions.

\noindent
\textbf{Attribute Recognition:} Attributes offer important mid-level cues of objects, usually in the form of a single word~\cite{fang2015captions, wu2017image}. Since we only use attributes for discerning the two images, we extend the notion of `single-word attribute' to a short phrase to enhance its discriminative power. 
For example, the attribute of ``next to building'' is actually frequent in everyday conversation and can be more expressive and discriminative than those single ``location'' attributes. To this end, we extract the commonly used $n$-gram expressions ($n\le3$) from region descriptions in Visual Genome dataset. We add the part-of-speech constraint to select for descriptive expressions.
An additional constraint is added so that the expressions should intersect with the top 1000 most frequent answers in the dataset. This helps filtering expressions that are less frequent or too specific.
Examples of expressions chosen to serve as our attributes include ``man'', ``stand'', ``in white shirt'', ``on wooden table'', ``next to tree''. More examples can be found in the supplementary material. The top $K=612$ constrained expressions are collected to form our attribute list $\{att_k\}$.

Next, we can associate each image region with its ground-truth attributes and train a visual attribute recognition model. We cast the learning as a multi-label classification problem. Specifically, we feed the visual representation $\mathbf{f}$ of each region into Multi-layer Perceptions (MLP) with a sigmoid layer to predict a K-d attribute score vector, $\mathbf{v}$. The MLP parameters are trained under a cross-entropy loss.

\noindent
\textbf{Attribute Selection:} Given the attribute score vectors $\mathbf{v}^A$, $\mathbf{v}^B \in \mathbb{R}^K$ extracted from two image regions $R^A$ and $R^B$, we want to choose an attribute pair $(att_i, att_j)$ that best distinguishes them. The chosen attributes should possess the following three desired properties: 

\noindent
1) Each attribute in the chosen pair should have highly contrasting responses on two regions. For examples, two regions with ``red'' and ``green'' attributes respectively would fulfill this requirement. 

\noindent
2) The chosen pair of attributes should be able to serve as a plausible answer for a single identical question. For instance, the ``red'' and ``green'' attributes both provide plausible answers to the question of ``What color is it?''.

\noindent
3) The chosen pair of attributes should be easily distinguished by visual observations. We define the visual dissimilarity as an intrinsic property of attributes independent to particular images.

We integrate these constraints into the following score function. Here we use a shorthand $(i,j)$ to represent $(att_{i},att_{j})$.
\begin{equation}\label{eq3}
\begin{aligned}
s(i,j) = &\underbrace{v_{i}^A (1-v_{i}^B) \cdot  v_{j}^B (1-v_{j}^A)}_{attribute \; score \; contrast} \\
&\cdot \underbrace{e^{\alpha s_q(i,j)}}_{question \; similarity} \cdot \underbrace{e^{-\beta s_f(i,j)}}_{visual \; dissimilarity},
\end{aligned}
\end{equation}
where $\alpha,\beta$ are the balancing weights among the three constraints, and $s_q(\cdot,\cdot)$ and $s_f(\cdot,\cdot)$ encode the question and feature similarities, respectively. We use the full score in Eq.~\eqref{eq3} to rank all $K^2$ attribute pairs in an efficient way, and select the top scoring pair to guide our VDQG. 
Next we explain each term in the score function:

\vspace{0.1cm}
\noindent
\underline{Attribute score contrast}. 
This term computes the score contrast of attributes between two image regions, where $v_i^A \in \mathbf{v}^A$ represents the score/response of $i$-th attribute on region $R^A$. Similar notational interpretation applies to other variables in this term. The score contrast of a discriminative attribute pair should be high.

\begin{figure}[t]
	\centering
	\includegraphics[width=1\linewidth]{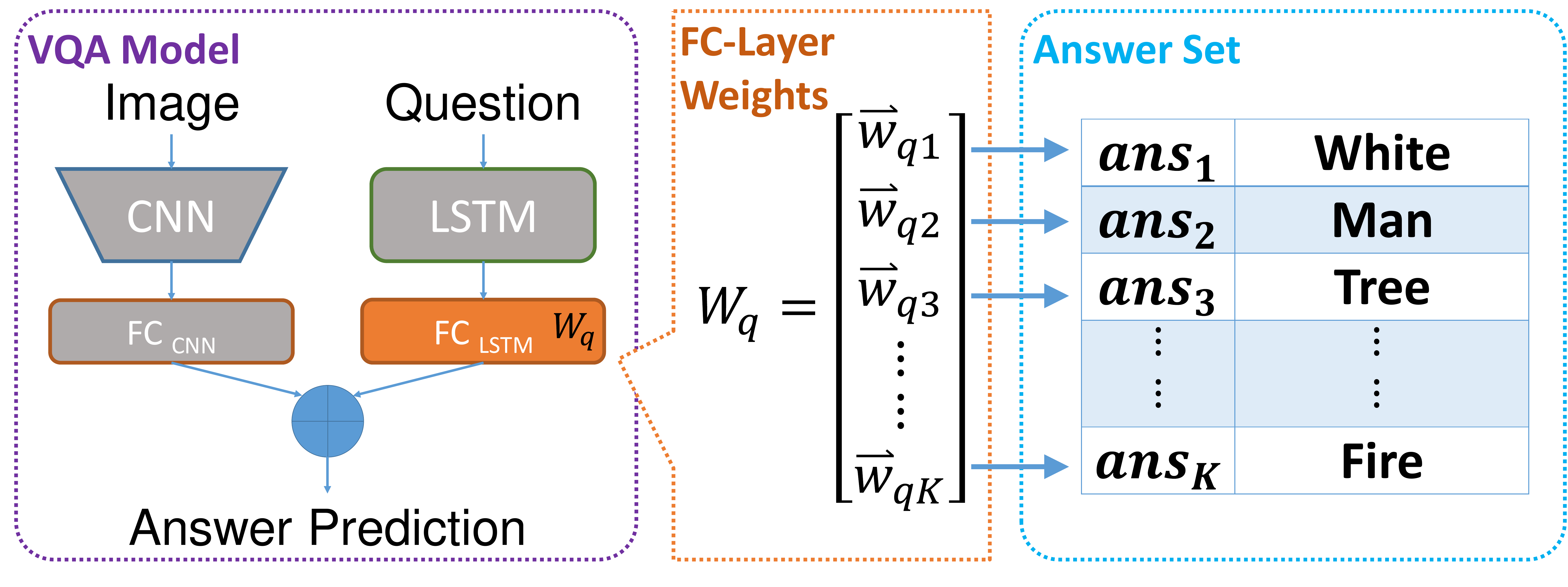}
	\vskip -0.2cm
	\caption{Question similarity scoring. We train a VQA model (left) using question-answer annotations of Visual Genome~\cite{krishna2017visual}. Since the answers overlap with our defined attributes, question similarity between two attributes $att_i$ and $att_j$ can be computed as the inner product of the corresponding $i$-th and $j$-th row vectors in the weight matrix of the $\mathrm{FC}_\mathrm{LSTM}$ layer.}
	\label{fig:vqa_model}
	\vskip -0.25cm
\end{figure}

\vspace{0.1cm}
\noindent
\underline{Question similarity $s_q(i,j)$}.
The question similarity score of a discriminative attribute pair should be large because they are intended to respond to the same identical question. Finding this similarity is non-trivial.
To compute the question similarity $s_q(i,j)$ of attributes $att_i$ and $att_j$, we train a small VQA model (see Fig.~\ref{fig:vqa_model}) that is capable of providing an answer given an input question and image. The model is trained using question-answer annotations from the Visual Genome dataset~\cite{krishna2017visual}. Note that we only train the model using question-answer annotations of which the answer is one of the attributes in $\{att_k\}$ that we define earlier (recall that our attribute set overlaps with the answer set). Thus the answer output of the VQA model is actually our attribute set and the model captures the question-attribute relations.

As illustrated in Fig.~\ref{fig:vqa_model}, the fully-connected layer after LSTM ($\mathrm{FC}_\mathrm{LSTM}$) contains a weight matrix $W_q$, of which the $i$-th row vector, denoted as $\overrightarrow{w}_{qi}$, is trained for prediction of attribute $att_i$.
In other words, this vector $\overrightarrow{w}_{qi}$ could serves as the representation of attribute $att_i$ in the question space.
Hence, the question similarity between $att_i$ and $att_j$ can be computed as the inner product of $\overrightarrow{w}_{qi}$ and $\overrightarrow{w}_{qj}$, denoted as $\langle \overrightarrow{w}_{qi}, \overrightarrow{w}_{qj}\rangle$.

\vspace{0.1cm}
\noindent
\underline{Visual similarity $s_f(i,j)$}.
The visual similarity score of a discriminative attribute pair should be small.
To determine the visual similarity $s_f(i,j)$ between attribute $att_i$ and $att_j$, we use the technique which we compute the question similarity.
Specifically, the fully-connected layer of our attribute recognition model contains a weight matrix $W_f$, of which the $i$-th row vector, denoted as $\overrightarrow{w}_{fi}$, is trained for prediction of attribute $att_i$. Consequently, the visual similarity between $att_i$ and $att_j$ can be computed as the inner product of $\overrightarrow{w}_{fi}$ and $\overrightarrow{w}_{fj}$, denoted as $\langle \overrightarrow{w}_{fi}, \overrightarrow{w}_{fj}\rangle$.

\subsection{CNN-LSTM with Attribute Conditions}
\label{sec:conditioned_lstm}

In this section, we describe the formulation of the attribute-conditioned LSTM. We start with a brief review of conventional CNN-LSTM.

\noindent
\textbf{Conventional CNN-LSTM}. 
In the typical CNN-LSTM language generation framework, CNN features $\mathbf{f}$ are first extracted from an input image. The features are then fed into the LSTM to generate language sequences. The model is trained by minimizing the negative log likelihood:
\begin{equation}\label{eq1}
\begin{aligned}
L &= \sum\nolimits_n -\log \; p(Q_n|\mathbf{f}_n)\\&= \sum\nolimits_n\sum\nolimits_t -\log \; p(q^n_t|q^n_{t-1,...,1},\mathbf{f}_n),
\end{aligned}
\end{equation}
where each question $Q_n$ comprises of a word sequence $\{q_t^n\}$.

\noindent
\textbf{Attribute-Conditioned LSTM}. To generate questions with specific intent, we utilize semantic attributes as an auxiliary input of the LSTM to condition the generation process. Ideally, when the model takes a ``red'' attribute, it would generate question like ``What is the color?''. 
We train such conditioned LSTM using the tuple $(\mathbf{f},Q,att_i)$, where $att_i$ is made out of the groundtruth answer of $Q$. Similar to Eq.~\eqref{eq1}, we minimize the negative log likelihood as follows:
\begin{equation}\label{eq5}
L = \sum\nolimits_n -log \; p(Q_n|\mathbf{f}_n,\sigma(att^n_{i})),
\end{equation}
where $\sigma(\cdot)$ is a feature embedding function for attribute input. We use Word2Vec~\cite{mikolov2013efficient} as the embedding function that can generalize across natural language answers and attributes.

Our goal is to generate one discriminative question collaboratively from two image regions $R^A$ and $R^B$ with the selected attribute pair $(att_i, att_j)$. Thus we duplicate the attribute-conditioned LSTM for each region and compute a joint question probability $p(Q|\mathbf{f}^A, \mathbf{f}^B, \sigma(att_i), \sigma(att_j))$, which can be expressed as
\begin{equation}\label{eq6}
\begin{aligned}
&p(q_t|q_{t-1,...,1},\mathbf{f}^A, \mathbf{f}^B, \sigma(att_i), \sigma(att_j)) = \\
&\!\!\!\!\!\! \frac{p(q_t|q_{t-1,...,1},\mathbf{f}_A, \sigma(att_i)) \cdot  p(q_t|q_{t-1,...,1},\mathbf{f}_B, \sigma(att_j))}{\sum_{q\in \mathcal{V}}{p(q|q_{t-1,...,1},\mathbf{f}_A, \sigma(att_i)) \cdot  p(q|q_{t-1,...,1},\mathbf{f}_B, \sigma(att_j))}},
\end{aligned}
\end{equation}
where $\mathcal{V}$ is the whole vocabulary. We use beam search to find the most probable questions according to Eq.~\eqref{eq6}.

\noindent
\textbf{Learning from Weak Supervision}.
As mentioned before, there are no public available paired-image datasets annotated with discriminative questions for fully-supervised learning.
Fortunately, due to the unique formulation of our approach, which extends CNN-LSTM to generate questions collaboratively from two image regions (see Eq.~\ref{eq6}), our method can be trained by just using `single image + single question' dataset.
We choose to utilize the rich information from Visual Genome dataset~\cite{krishna2017visual}. In particular, we extract 1445k image-related question-answer pairs and their grounding information,~\ie,~region bounding box. We randomly split the question-answer pairs into training (70\%), validation (15\%) and testing (15\%) sets, where questions referring to the same image will only appear in the same set. We also utilize the associated region descriptions to enrich the textual information for our attribute-conditioned model (Sec.~\ref{sec:attribute_model}). 
It is worth noting that the training and validation sets are only used for our model training in a weakly-supervised manner, while the testing set is used to construct the VDQG dataset as introduced in Sec.~\ref{sec:dataset}.

\section{Experiments}
\label{sec:experiments}

\vspace{0.1cm}
\noindent
\textbf{Methods}. We perform experiments on the proposed VDQG datasets and evaluate the following methods:

\noindent
1) \textit{Our Approach} (ACQG).
We call our approach as Attribute-Conditioned Question Generation (ACQG). We establish a few variants based on the way discriminative attributes are selected. ACQG$_{ac}$ only uses the attribute score contrast in Eq.~\eqref{eq3}. 
ACQG$_{ac+qs}$ uses both attribute score contrast and question similarity. Lastly, ACQG$_{full}$ uses all the terms for attribute selection. For each sample,we select top-5 attribute pairs and generate questions for each pair. The final output is the question with the highest score, which is the product of its attribute score (Eq.~\ref{eq3}) and question probability (Eq.~\ref{eq6}). This achieves a better performance than only using top-1 attribute pair.

\noindent
2) \textit{CNN-LSTM}.
We modify the state-of-the-art image captioning CNN-LSTM model~\cite{donahue2015long} for the VDQG task. Specifically, we adopt Inception-ResNet~\cite{szegedy2016inception} as the CNN part, followed by two stacked 512-d LSTMs. We also extend the framework to accommodate image pair input following Eq.~\eqref{eq6} without using pair attributes as the condition.

\noindent
3) \textit{Retrieval-based Approach} (Retrieval).
It is shown in~\cite{mostafazadeh-EtAl:2016:P16-1} that carefully designed retrieval approaches can be competitive with generative approaches for their VQG task. 
Inspired by~\cite{mostafazadeh-EtAl:2016:P16-1}, we prepare a retrieval-based baseline for the VDQG task.
Our training set consists of questions annotated on image regions. Given a test image pair, we first search for the $k$ nearest neighbor ($k=100$) training image regions for the pair, and use the training questions annotated on these retrieved regions to build a candidate pool. 
For each question in the candidate pool, we compute its similarity to the other questions using BLEU~\cite{papineni2002bleu} score. The candidate question with the highest score will be associated with the input image pair.

\vspace{0.1cm}
\noindent
\textbf{Evaluation Metrics}.
To evaluate a generated question, we hope to reward a match with the positive ground-truth questions, and punish a match with the negative ground-truth questions. 
To this end, we use $\Delta$BLEU~\cite{galley2015deltableu} as our main evaluation metric, which is tailored for text generation tasks that admit a diverse range of possible outputs. 
Mostafazadeh \etal~\cite{mostafazadeh-EtAl:2016:P16-1} show that $\Delta$BLEU has a strong correlation with human judgments in visual question generation task. In particular, given a reference (annotated question) set $\{r_{i,j}\}$ and the hypothesis (generated question) set$\{h_i\}$, where $i$ is the sample index and $j$ is the annotated question index of $i$-th sample, $\Delta$BLEU score is computed as:
\begin{equation}\label{eq7}
\begin{aligned}
\Delta\mathrm{BLEU} = \mathrm{BP}\cdot \exp(\sum\nolimits_n{\log p_n})
\end{aligned}
\end{equation}
The corpus-level $n$-gram precision is defined as:
\begin{equation}
\begin{aligned}
p_n = {{\sum_i\sum_{g\in \mathrm{n-grams}(h_i)} \max_{j:g\in r_{i,j}}\{w_{i,j}\cdot\#_g(h_i,r_{i,j})\}}\over{\sum_i\sum_{g\in \mathrm{n-grams}(h_i)}\max_j \{w_{i,j}\cdot\#_g(h_i)\}}},
\end{aligned}
\end{equation}
where $\#_g(\cdot)$ is the number of occurrences of n-gram $g$ in a given question, and $\#_g(u,v)$ is the shorthand for $\min\{\#_g(u),\#_g(v)\}$. And the brevity penalty coefficient $\mathrm{BP}$ is defined as:
\begin{equation}
\begin{aligned}
\mathrm{BP}=\left\{ 
\begin{array}{ll}
1& \text{if}~ \rho > \eta \\ 
e^{1-\eta/\rho}&\text{if}~ \rho \le \eta
\end{array}%
\right.
\end{aligned},
\end{equation}
where $\rho$ and $\eta$ are respectively the length of generated question and effective annotation length. We respectively set the score coefficients of strong-positive samples, weak-positive samples and negative samples to be 1.0, 0.5 and -0.5. We use a equal weights for up to 4-grams.

As a supplement, we also use BLEU~\cite{papineni2002bleu} and METEOR~\cite{lavie2014meteor} to evaluate the textual similarity between generated questions and positive annotations in the test set.

\subsection{Results}

We conducted two experiments based on the VDQG dataset. The first experiment was conducted on the full samples. The second experiment was performed by using only a hard subset of VDQG. We constructed the hard subset by selecting 50\% samples with a lower ratio of positive annotations within each object category.

Table~\ref{table:result_full} summarizes the results on the full VDQG dataset. The proposed method outperforms baseline methods according to all metrics. 
We also performed ablation study by gradually dropping the similarity terms in Eq.~\eqref{eq3} out of our full model. The results suggest that question similarity dominates the performance improvement while other terms also play an essential role. It is noted that ACQG$_{ac}$ yields poor results in comparison to the baseline CNN-LSTM. Based on our conjecture, the attribute score contrast term may be too simple therefore overwhelmed by the noisy prediction scores of attributes.
Experimental results on the hard subset are shown in Table~\ref{table:result_hard}. Compared with the results in Table~\ref{table:result_full}, the performance gap between ACQG$_{full}$ and non-attribute-guided models increases in hard cases, which shows the significance of discriminative attributes in the task of VDQG.

\begin{table}[t]
    \centering
    \caption{Experiment results on full VDQG dataset.}
    \vskip -0.2cm
    \label{table:result_full}
    \small{
        \begin{tabular}{l c c c}
            \hline
            \textbf{Model} & $\Delta$BLEU & BLEU & METEOR \\
            \hline
            Human$_{top}$ &69.2&85.5&57.5\\
            Human$_{random}$ &62.9&82.4&54.9\\
            \hline
            Retrieval &24.3&42.5&29.1\\
            CNN-LSTM &33.4&56.2&37.3\\
            \hline
            ACQG$_{ac}$&29.4&52.9&35.3\\
            ACQG$_{ac+qs}$&40.1&59.1&39.6\\
            ACQG$_{full}$ &\textbf{40.6}&\textbf{59.4}&\textbf{39.7}\\                
            \hline
        \end{tabular}
    }
\end{table}

\begin{table}[t]
    \centering
    \caption{Experiment results on VDQG hard subset.}
    \vskip -0.2cm
    \label{table:result_hard}
    \small{
        \begin{tabular}{l c c c}
            \hline
            \textbf{Model} & $\Delta$BLEU & BLEU & METEOR \\
            \hline
            Human$_{top}$ &62.3&79.2&52.2\\
            Human$_{random}$ &53.7&74.9&48.9\\
            \hline
            Retrieval &13.4&36.9&25.9\\
            CNN-LSTM &20.3&47.8&32.7\\
            \hline
            ACQG$_{ac}$&13.5&44.3&30.4\\
            ACQG$_{ac+qs}$&32.6&53.2&36.1\\
            ACQG$_{full}$ &\textbf{33.5}&\textbf{53.6}&\textbf{36.4}\\                
            \hline
        \end{tabular}
    }
    \vspace{-0.2cm}
\end{table}

We also performed an interesting experiment based on the collected question annotations in VDQG dataset. Specifically, `Human$_{top}$' indicates the first-annotated positive question of each sample, while `Human$_{random}$' indicates a random positive annotation among all the human annotations of each sample. It is reasonable to assume that the first-written questions are likely to ask the most distinguishing differences between two images. From both Tables~\ref{table:result_full} and \ref{table:result_hard}, we observe that `Human$_{top}$' consistently outperforms `Human$_{random}$. The results suggest the effectiveness of the proposed VDQG dataset and metric settings for VDQG evaluation.

\subsection{User Study}

\begin{figure}[t]
    \centering
    \includegraphics[width=\linewidth]{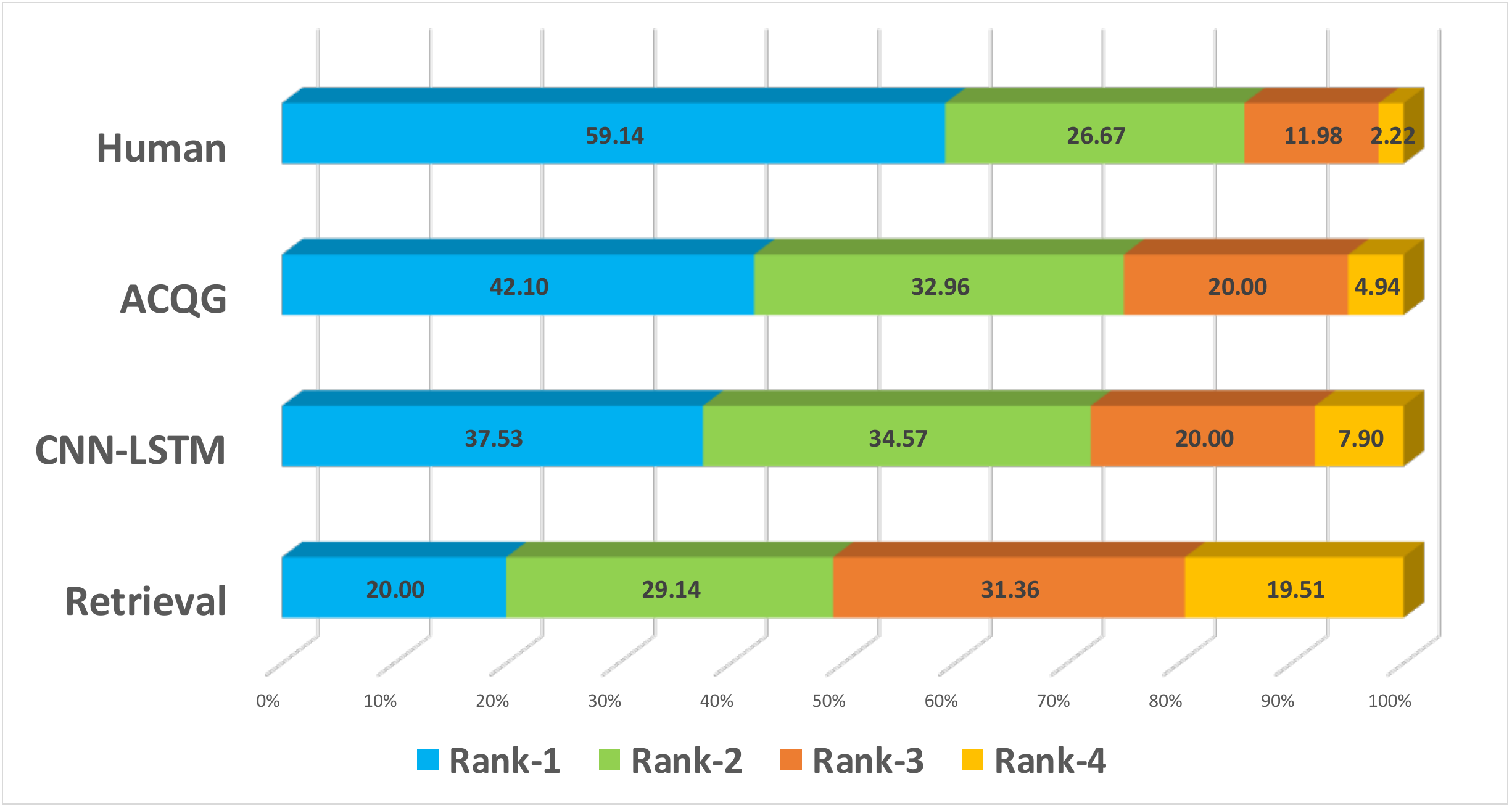}
    \vskip -0.25cm
    \caption{User study on VDQG full.}
    \label{fig:userstudy_question_all}
\end{figure}

\begin{figure}[t]
    \centering
    \includegraphics[width=\linewidth]{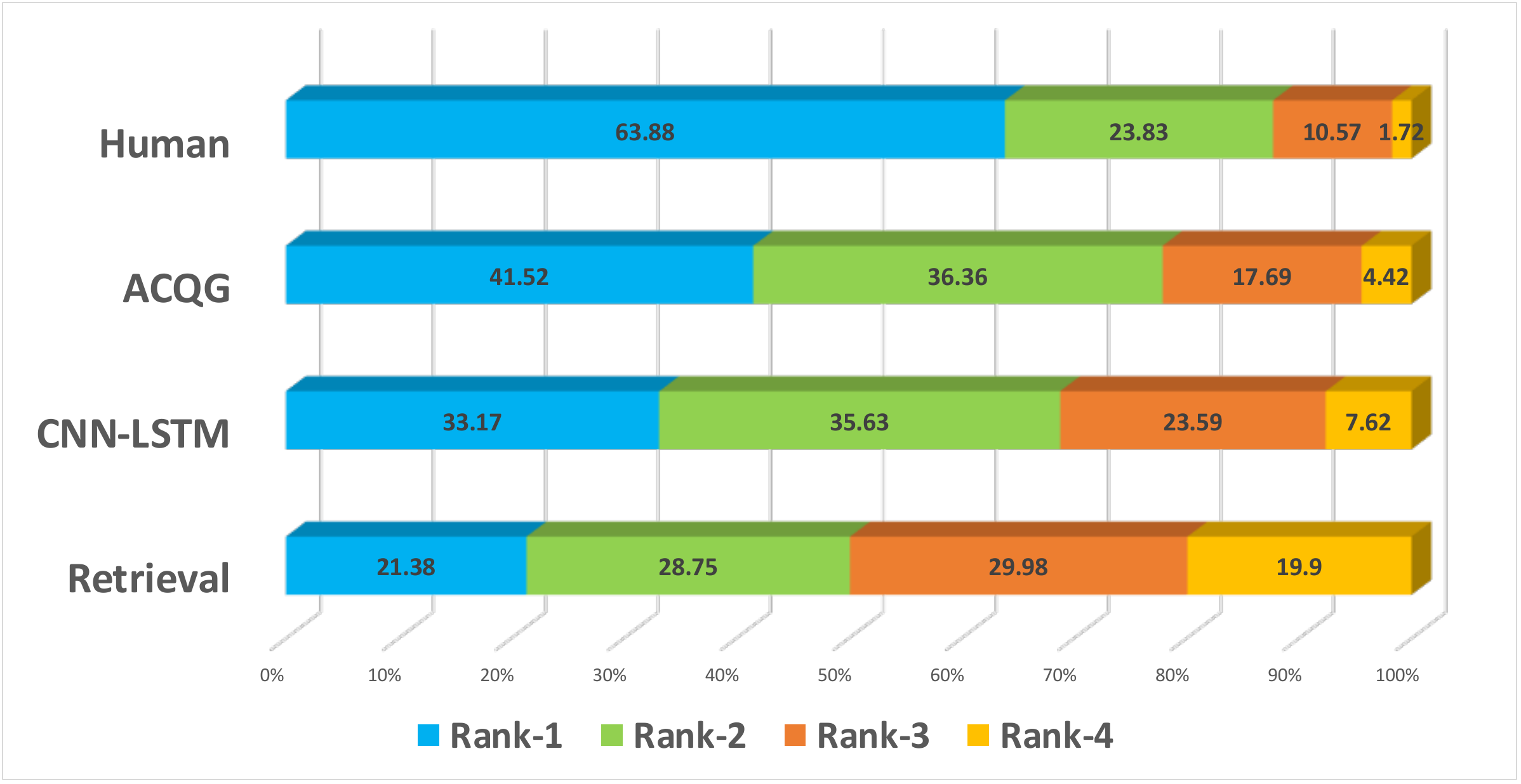}
    \vskip -0.25cm
    \caption{User study on VDQG hard subset.}
    \label{fig:userstudy_question_hard}
    \vskip -0.25cm
\end{figure}

We gathered a total of 27 participants to join our user study. Each time we showed the participant an image pair and four questions generated respectively by a human annotator (the groundtruth), the proposed ACQG$_{full}$, CNN-LSTM, and Retrieval. Then the participant was asked to rank these questions according to their capability of distinguishing the given image pair. Figure~\ref{fig:userstudy_question_all} shows the results of user study. We also separately analyze the hard samples, and show the results in Fig.~\ref{fig:userstudy_question_hard}.
The proposed ACQG$_{full}$ outperforms other baseline models in the user study. It is observed that the performance gap becomes more significant on hard samples.

\section{Comparison with Referring Expression}
A referring expression is a kind of unambiguous description that refers to a particular object within an image. Despite the linguistic form differences between the discriminative question and the referring expression, they have the common objective of disambiguation. In this section, we compared discriminative question with referring expression by conducting a user study with 14 participants. Specifically, each time we showed the participant an image with two ambiguous objects marked with their respective bounding boxes. Meanwhile, we showed the participant a referring expression\footnote{We generate referring expressions using the state-of-the-art REG model~\cite{mao2016generation} trained on RefCOCO+ dataset~\cite{yu2016modeling}. The images used in the user study are selected from the validation set of RefCOCO+. In particular, we select the images containing two ambiguous objects.} or a discriminative question with its conditioning attribute that refers to one of the objects. Then the participant was asked to retrieve the referred object by the given information. We compute the mean retrieval accuracy to measure the disambiguation capability of the given textual information. 

\begin{figure}[t]
    \centering
    \includegraphics[width=\linewidth]{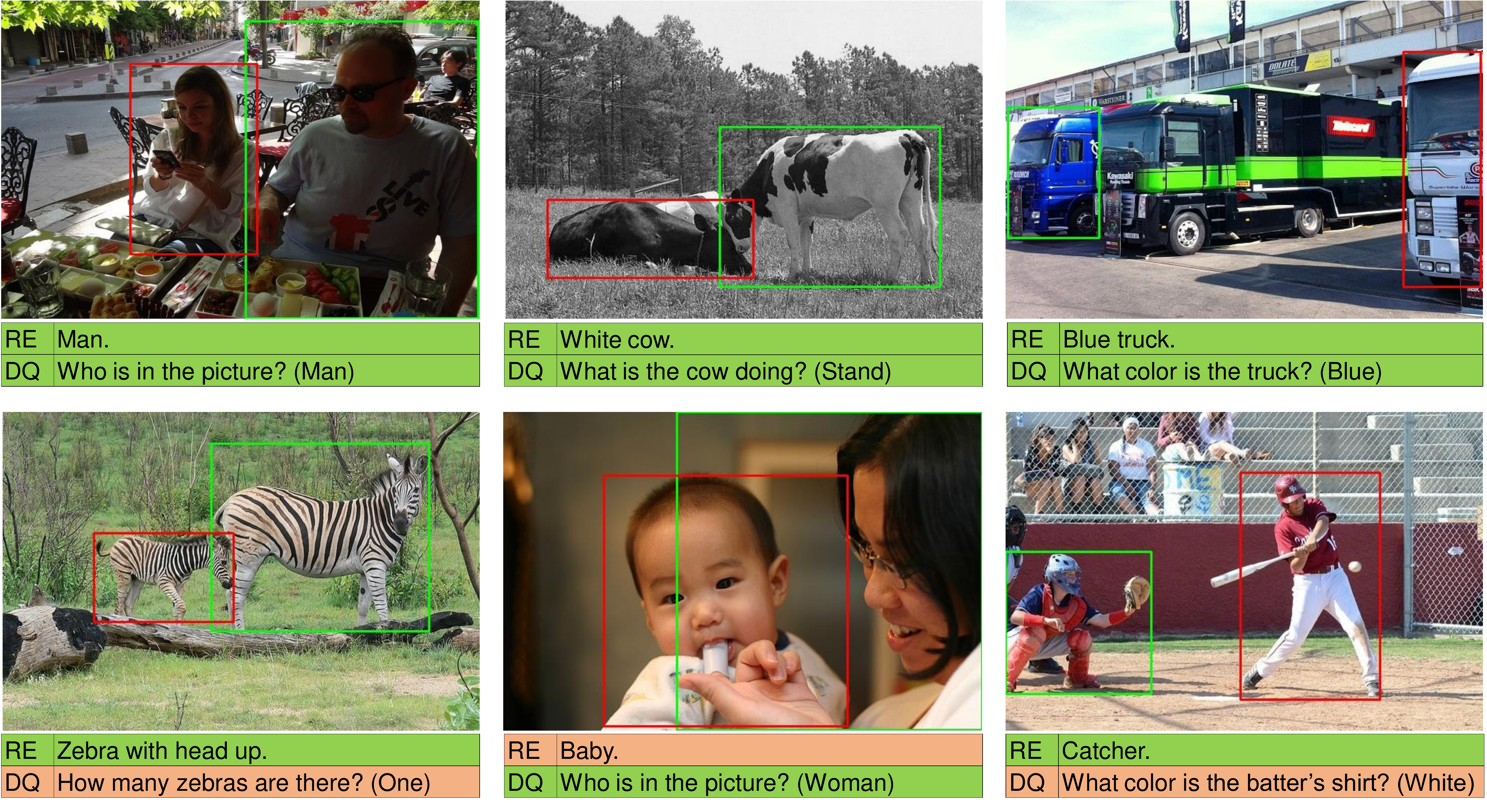}
    \vskip -0.25cm
    \caption{Visualization of the Discriminative Question (DQ) and Referring Expression (RE) generated from the ambiguous objects in images. Referred objects and distractors are marked with green and red bounding boxes respectively. The second row shows some failure cases.}
    \label{fig:compare}
    \vskip -0.25cm
\end{figure}

The results are interesting -- showing referring expressions results in a mean retrieval accuracy of $65.14\%$,Πwhile showing discriminative question$+$attribute achieves a competitive result of $69.51\%$. In Fig.~\ref{fig:compare} we show some of the generated referring expressions and discriminative questions on ambiguous objects within images. It is interesting to notice that referring expressions and discriminative questions fail in different cases, which indicates that they could be further studied as complementary approaches to visual disambiguation.

\section{Conclusion}
\label{sec:conclusion}

We have presented a novel problem of generating discriminative questions to help disambiguate visual instances. We built a large-scale dataset to facilitate the evaluation of this task. Besides, we proposed a question generation model that is conditioned on discriminative attributes. The method can be trained by using weak supervisions extracted from existing VQA dataset (single image + single question), without using full supervision that consists of paired-image samples annotated with discriminative questions.

\vspace{0.1cm} \noindent
\textbf{Acknowledgement}: This work is supported by SenseTime Group Limited and the General Research Fund sponsored by the Research Grants Council of the Hong Kong SAR (CUHK 416713, 14241716, 14224316. 14209217).

{\small
	\bibliographystyle{ieee}
	\bibliography{vqg,submission}
}

\newpage
\section*{Supplementary Material}

\subsection*{A. VDQG Dataset}

\begin{figure*}[t]
	\centering
	\includegraphics[width=\linewidth]{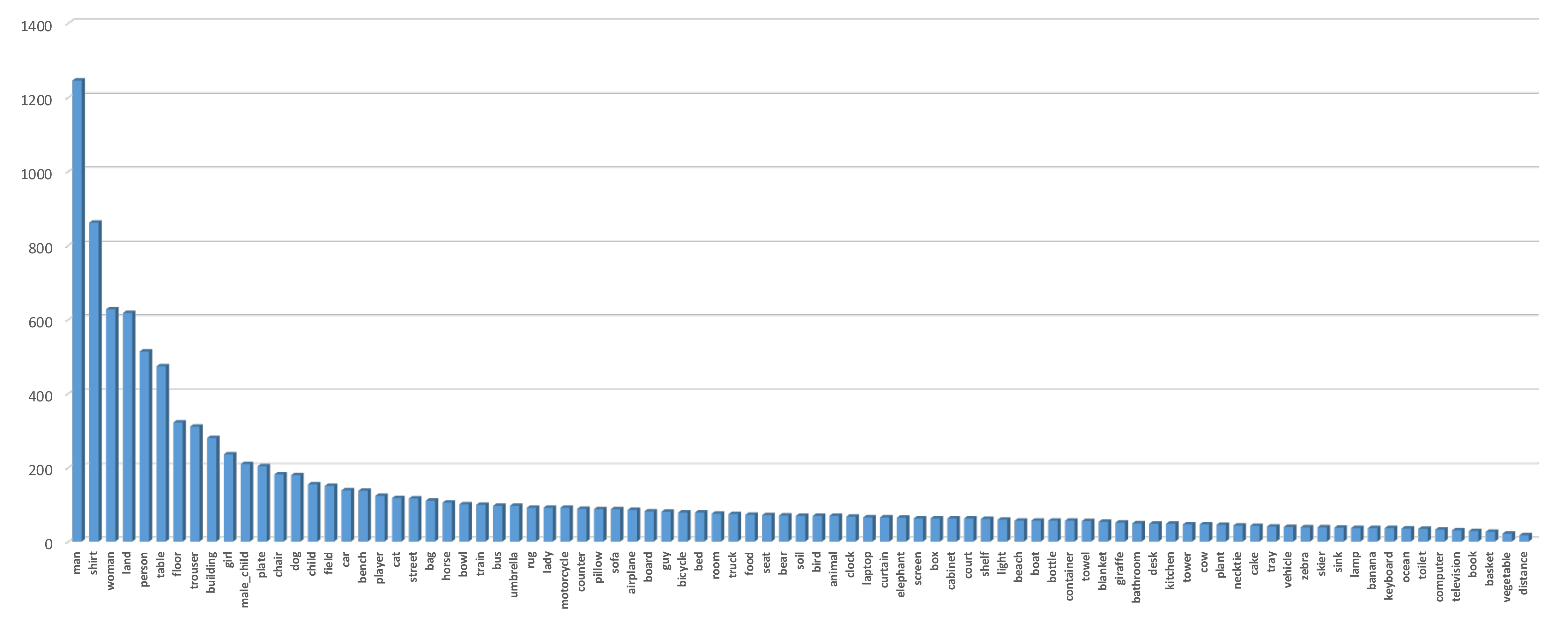}
	\vskip -0.25cm
	\caption{Object category distribution of VDQG dataset.}
	\label{fig:object}
\end{figure*}

\noindent
\textbf{Object Category.}
We selected 87 object categories from the annotation of Visual Genome datasets~\cite{krishna2017visual} to construct the VDQG dataset. Figure~\ref{fig:object} shows the list of object category and the number of samples belonging to each object category.

\begin{figure*}[t]
	\centering
	\begin{subfigure}[t]{0.48\textwidth}
		\includegraphics[width=\textwidth]{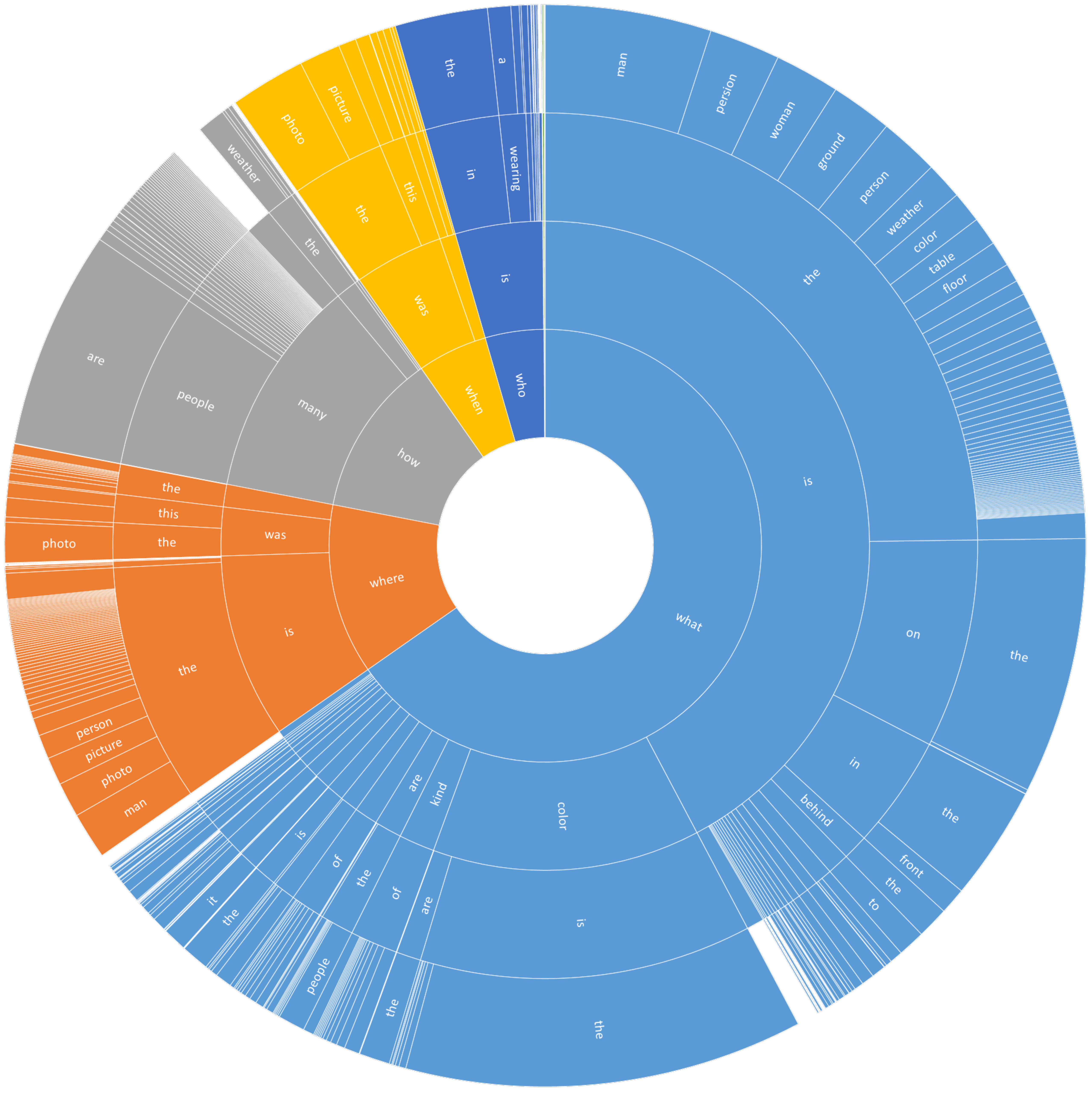}
		\caption{VDQG}
		\vskip -0.25cm
	\end{subfigure}
	~
	\begin{subfigure}[t]{0.48\textwidth}
		\includegraphics[width=\textwidth]{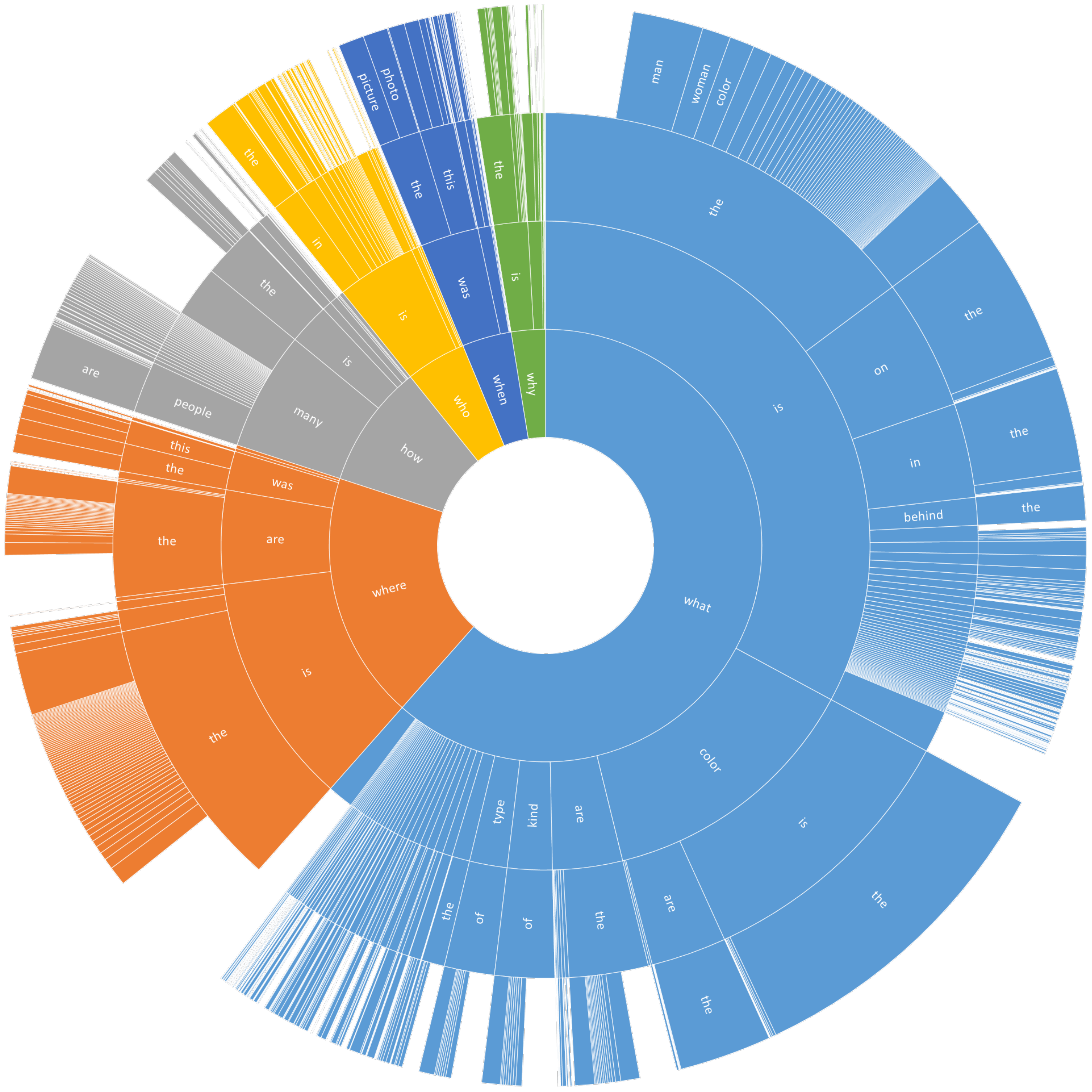}
		\caption{Visual Genome}
		\vskip -0.25cm
	\end{subfigure}
	\caption{$N$-gram sequence distribution of VDQG dataset (a) and Visual Genome dataset (b).}
	\label{fig:ngram}
\end{figure*}

\noindent
\textbf{Question Type.}
Figure~\ref{fig:ngram} visualizes the most frequent $n$-gram ($n \le 4$) sequences of questions in the VDQG dataset as well as the Visual Genome dataset. We observe that the question type distributions of these two datasets are similar to each other. A significant difference is that there is almost no ``why'' type question in VDQG dataset, which is reasonable because this type of question is hardly used to distinguish similar objects.

\begin{figure*}[t]
	\centering
	\includegraphics[width=\linewidth]{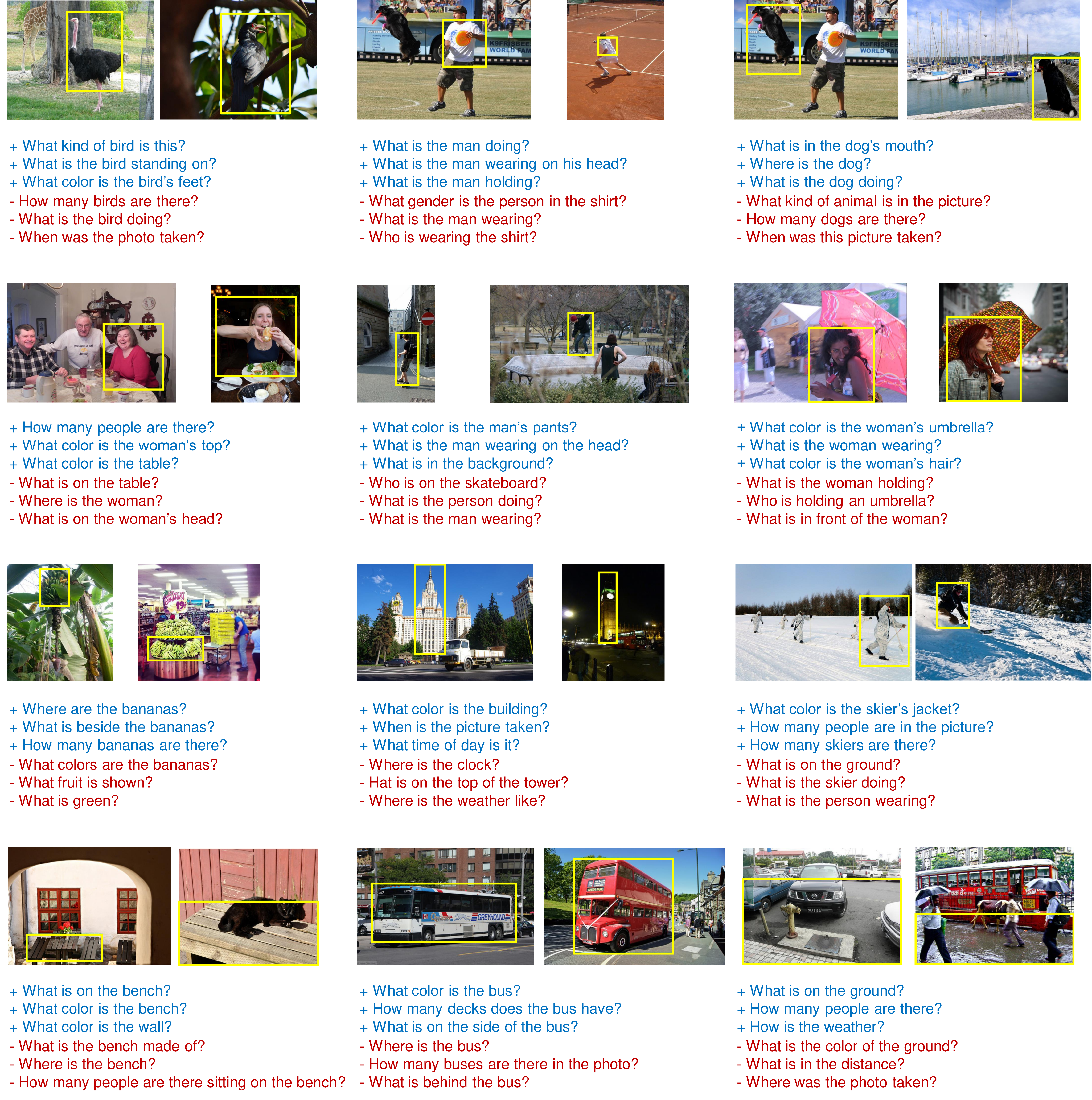}
	\vskip -0.25cm
	\caption{Example of image pairs and the associated positive and negative question annotations in the proposed VDQG dataset. Positive and negative questions are written in blue and red, respectively.}
	\label{fig:dataset_examples_supp}
\end{figure*}

\noindent
\textbf{Examples.}
We show some examples of the VDQG dataset in Fig.~\ref{fig:dataset_examples_supp}.

\subsection*{B. Implementation Details}

\noindent
\textbf{Attributes.}
We built an attribute set by extracting the commonly used $n$-gram expressions ($n\le3$) from region descriptions available in the Visual Genome dataset. And the part-of-speech constraint has been taken into consideration to select for discriminative expressions. Table~\ref{table:attribute} shows the part-of-speech constraints we use and the most frequent attributes.

\begin{table}[h]
	\centering
	\caption{Part-of-speech constraint on $n$-gram expressions to extract attributes}
	\label{table:attribute}
	\small{
		\begin{tabular}{c|c}
			\hline
			Part of speech & Top attributes \\
			\hline\hline
			$<$NN$>$ & man, woman, table, shirt, person \\
			$<$JJ$>$ & white, black, blue, brown, green \\
			$<$VB$>$ & wear, stand, hold, sit, look \\
			$<$CD$>$ & one, more\_than\_one \tablefootnote{We merge all numbers that are greater than one into one label ``more\_than\_one''.} \\
			$<$JJ,NN$>$ & white plate, teddy bear, young man\\
			$<$VB,NN$>$ & play tennis, hit ball, eat grass\\
			$<$IN,NN$>$ & on table, in front, on top, in background\\
			$<$NN,NN$>$ & tennis player, stop sign, tennis court\\
			$<$VB,NN,NN$>$ & play video game\\
			$<$IN,NN,NN$>$ & on tennis court, on train track\\
			\hline
		\end{tabular}
	}
\end{table}

\noindent
\textbf{Model Optimization.}
We implement our model using Caffe~\cite{jia2014caffe} and optimize the model parameters using Adam~\cite{kingma2015adam} algorithm. For the attribute recognition model, we use a batchsize of 50 and train for 100 epochs. For the attribute-conditioned LSTM model, we use a batchsize of 50 and train for 30 epochs, where gradient clipping is applied for stability. The parameters of CNN network has been pre-trained on ImageNet~\cite{russakovsky2015imagenet}, and fixed during finetuning for efficiency.

\subsection*{C. Qualitative Results} 

\begin{figure*}[t]
	\centering
	\includegraphics[width=\linewidth]{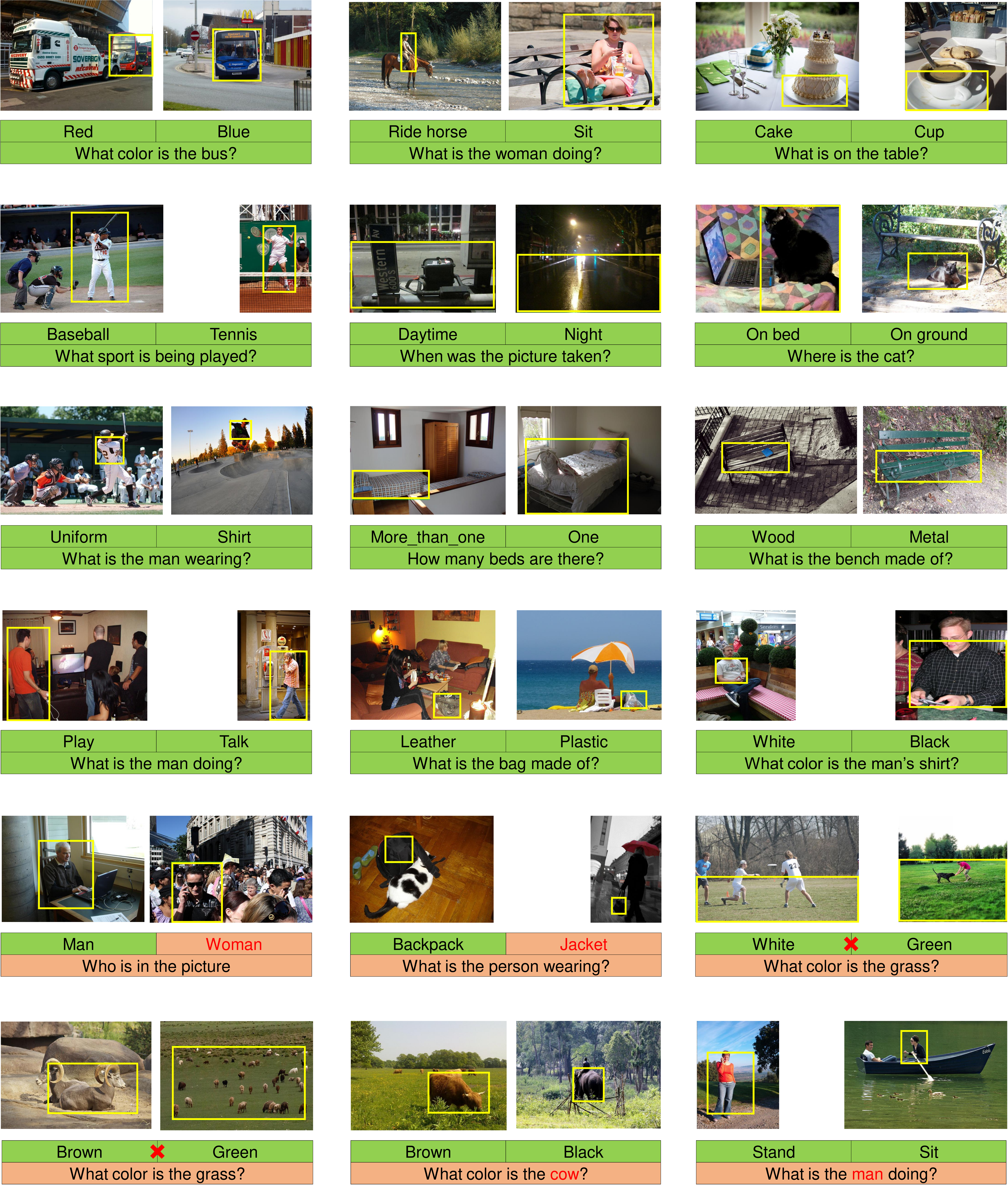}
	\vskip -0.25cm
	\caption{Discriminative questions generated by our approach. Under each ambiguous pair, the first line shows the distinguishing attribute pair selected by the attribute model, and the second line shows the questions generated by the attibute-conditioned LSTM. The last two rows show some failure cases.}
	\label{fig:results}
\end{figure*}

Figure~\ref{fig:results} shows some examples of discriminative question generated using our approach. The experimental result shows that our model is capable of capturing distinguishing attributes and generate discriminative questions based on the attributes. Some failure cases are shown at the last two rows in Figure~\ref{fig:results}. We observe that the failure cases are caused by different reasons. Specifically, the first two failure examples are caused by incorrect attribute recognition; the following two failure examples are caused by pairing attributes of different type of objects (\eg~pairing ``green'' of the grass and ``white'' of the people's clothes); and the last two failure examples are caused by incorrect language generation.

\end{document}